\documentclass[runningheads]{llncs}

\usepackage[year=2024]{eccv}

\usepackage{eccvabbrv}

\usepackage{graphicx}
\usepackage{booktabs}
\usepackage{caption}
\usepackage{algorithm, algorithmic}
\usepackage{multirow}
\usepackage{wrapfig}
\usepackage[accsupp]{axessibility}  

\usepackage[pagebackref,breaklinks,colorlinks,citecolor=eccvblue]{hyperref}

\usepackage{orcidlink}

\begin{document}

\title{Model Will Tell: Training Membership Inference for Diffusion Models} 


\author{
Xiaomeng Fu\inst{1,2} \and
Xi Wang\inst{1}\thanks{Corresponding Author.} \and
Qiao Li\inst{1} \and
Jin Liu\inst{1,2} \and
Jiao Dai\inst{1} \and
Jizhong Han\inst{1}
}

\authorrunning{Fu et al.}

\institute{Institute of Information Engineering, Chinese Academy of Sciences \and
School of Cyber Security, University of Chinese Academy of Sciences\\ 
\email{\{fuxiaomeng, wangxi1, liqiao, liujin, daijiao, hanjizhong\}}@iie.ac.cn
}

\maketitle

\begin{abstract}
Diffusion models pose risks of privacy breaches and copyright disputes, primarily stemming from the potential utilization of unauthorized data during the training phase. The Training Membership Inference (TMI) task aims to determine whether a specific sample has been used in the training process of a target model, representing a critical tool for privacy violation verification. However, the increased stochasticity inherent in diffusion renders traditional shadow-model-based or metric-based methods ineffective when applied to diffusion models. Moreover, existing methods only yield binary classification labels which lack necessary comprehensibility in practical applications. 
In this paper, we explore a novel perspective for the TMI task by leveraging the intrinsic generative priors within the diffusion model. Compared with unseen samples, training samples exhibit stronger generative priors within the diffusion model, enabling the successful reconstruction of substantially degraded training images.
Consequently, we propose the \textbf{D}egrade \textbf{R}estore \textbf{C}ompare (\textbf{DRC}) framework. In this framework, an image undergoes sequential degradation and restoration, and its membership is determined by comparing it with the restored counterpart. 
Experimental results verify that our approach not only significantly outperforms existing methods in terms of accuracy but also provides comprehensible decision criteria, offering evidence for potential privacy violations.
\end{abstract}

\section{Introduction}
Recently, the emergence of large models, particularly the Diffusion Model~\cite{ho2020denoising,song2020score}, has significantly enhanced the authenticity and practicality of Artificial Intelligence Generated Content (AIGC). However, these data-intensive models are predominantly trained on extensive unauthorized data~\cite{schuhmann2021laion,schuhmann2022laion} scrabbled from the internet, thereby ignoring the proprietary rights and privacy of the original owners. Consequently, this has amplified the risk of privacy breaches and copyright disputes for users~\cite{lee2023stable,edwards2022artists,brittain2023getty}. For example, Getty Images has filed a lawsuit against Stability AI recently, accusing that the company utilized 12 million Getty Images to train its model without necessary permissions~\cite{brittain2023getty}. Furthermore, the improper use of personal identity to generate facial images also poses a significant concern in the realm of privacy protection. To address the aforementioned challenges, a pivotal approach is to ensure that users are informed whether their private data is being utilized for model training.

\begin{figure*}[ht!]
\centering
\begin{subfigure}{.45\linewidth}
  \includegraphics[width=\linewidth]{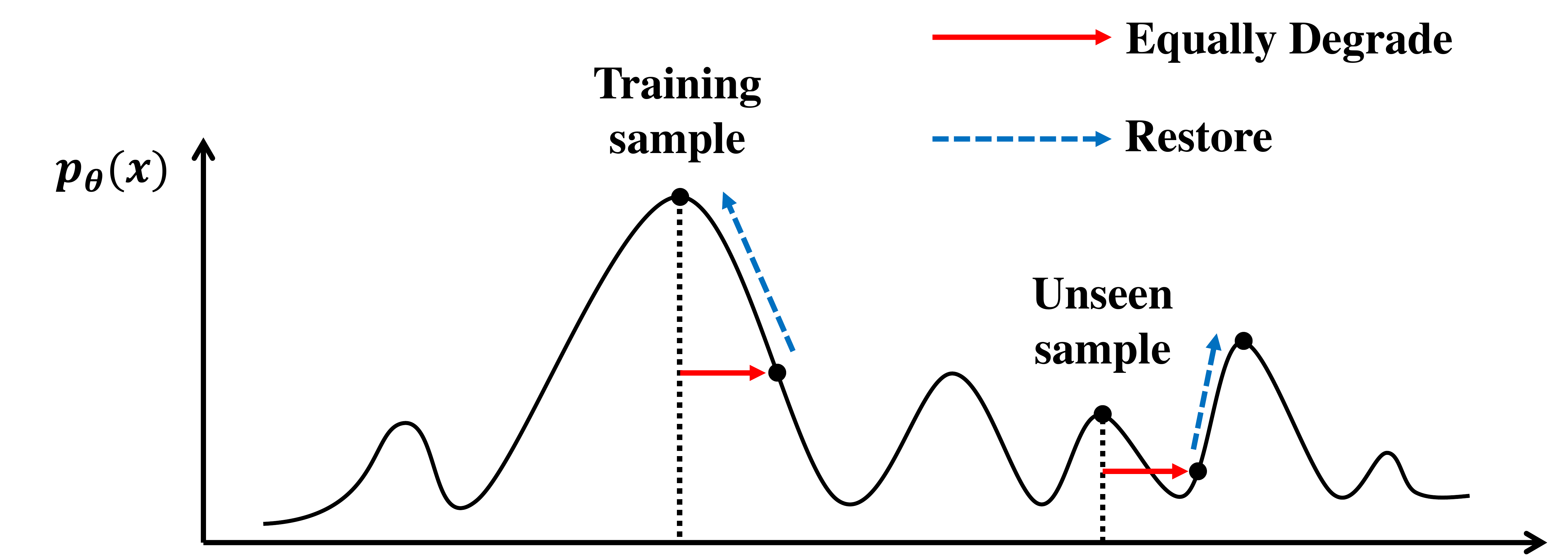}
  \caption{The intuition of DRC.}
  \label{fig:cifar-10}
\end{subfigure}
\begin{subfigure}{.50\linewidth}
  \includegraphics[width=\linewidth]{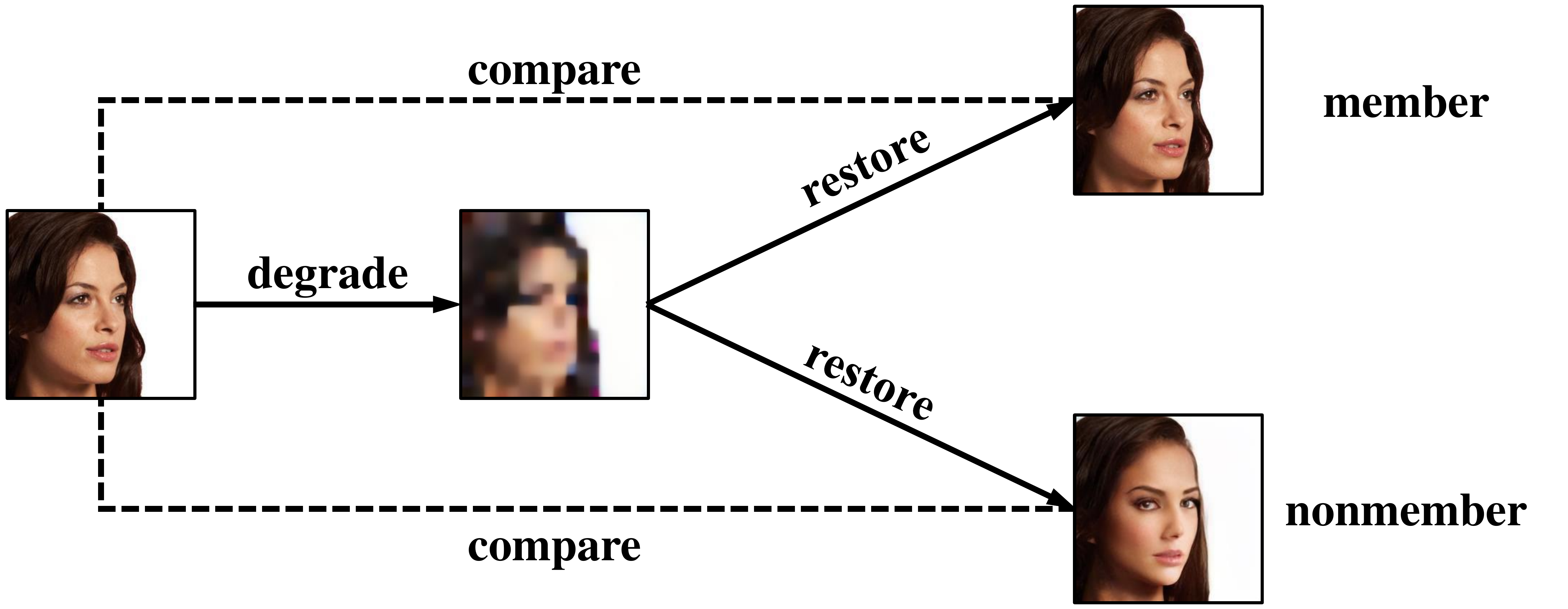}
  \caption{The pipeline of DRC.}
  \label{fig:cifar-100}
\end{subfigure}

\caption{(a) The intuition of DRC. A training sample will exhibit a more pronounced peak compared with an unseen sample. The training sample and the unseen sample are equally degraded. The degraded training sample can be restored to its original peak, while the capability is not shared by the unseen sample. (b) The pipeline of DRC. The original image is first intentionally degraded, and then restored by the diffusion model. Finally, a comparison is made between the original image and its restored counterpart to determine whether the original image is a training member.}
\label{motivation}
\end{figure*}

We resort to Training Membership Inference (TMI) tasks to empower users with the ability to discern potential threats to their private data. The TMI task, also denoted as Membership Inference Attacks (MIAs)~\cite{hu2022membership}, is aimed to identify whether a data sample has been used in the training of a machine learning model. Historically, TMI methods~\cite{shokri2017membership,hui2021practical,leino2020stolen,long2020pragmatic,salem2019ml,truex2019demystifying} have been primarily developed for classification models. Some researchers~\cite{chen2020gan,hayes2019logan,hilprecht2019monte} have expanded these techniques to encompass generative methods, such as Generative Adversarial Networks (GANs)~\cite{goodfellow2014generative} and Variational Autoencoders (VAEs)~\cite{kingmaauto}. However, the applicability of existing TMI methods to diffusion models is limited. Specifically, one prevalent approach in TMI involves training a shadow model to mimic the target model’s behavior. This technique, however, is not suitable for diffusion models, as it is computationally intensive to train a shadow model with comparable parameters. Another category of TMI methods is metric-based, where a metric (typically the loss value) is selected as a surrogate for the model’s response towards each sample. The membership of the sample is then determined based on the numerical value of the metric. These methods encounter difficulties when dealing with the inherent randomness of the generation process in diffusion models. Moreover, these methods perceive TMI as a binary classification task, providing a label without any explanation. The label-only classification results obtained from these methods are not trustworthy for end users without some level of technical knowledge, thereby impeding the practical application in the real world. This highlights the need for more robust and user-friendly TMI methods for diffusion models.

In this paper, we explore a novel perspective for the TMI task. Large models inherently possess powerful capabilities for understanding and reasoning. Taking inspiration from GPT's self-detection of its generated language~\cite{ouyang2022training,radford2019language}, we attempt to leverage the inherent generative capability of the diffusion model to accomplish the TMI task.  
As illustrated in Figure~\ref{motivation}(a), the training process of the diffusion model is self-referential, with the training image serving as the objective. Consequently, the sample from the training set will manifest as a more pronounced peak compared to an unseen sample. By strategically degrading a training image in a controlled manner, while ensuring its proximity to the peak, there is a high probability that the diffusion model can restore it back to its original peak. However, when an equivalent level of degradation is applied to an unseen image, the diffusion model may restore it to an alternative peak, obtaining an entirely different image.

Therefore, we design a novel TMI framework, the \textbf{D}egrade \textbf{R}estore \textbf{C}ompare (\textbf{DRC}), to identify the training membership of a specific image. As depicted in Figure~\ref{motivation}(b), our approach involves a three-step process. First, the original image undergoes intentional degradation. Then, the generative priors of the target diffusion model are employed to restore the degraded image. Finally, a comparison is made between the restored image and the original image. The semantic distance between the original image and its restored counterpart is utilized to determine whether the original image is a training member. In the scenario where the original image is a training sample, the degraded image can be restored to an image almost identical to the original image. Conversely, an unseen sample may converge to an image entirely distinct from the original.
Our experimental results demonstrate that our method achieves state-of-the-art performance. Besides, the underlying mechanism of our proposed method is simple and comprehensible for end users with limited technical knowledge about computer vision, thereby enhancing the accessibility and applicability of our research.

In summary, this paper offers the following significant contributions: 
\begin{itemize}
    \item \textbf{Innovative TMI framework.}
    We introduce the \textbf{D}egrade \textbf{R}estore \textbf{C}ompare framework, a novel approach that capitalizes on the inherent reasoning and generative capabilities of the diffusion model for TMI tasks.
    \item \textbf{Comprehensibility.}
    The fundamental mechanism of our proposed method is intuitive and comprehensible, enhancing the accessibility and applicability of our research. 
    \item \textbf{Accuracy.} Experiments show that our proposed method outperforms contemporary TMI methods by a large margin, demonstrating its effectiveness. 
    
\end{itemize}

\section{Related Work and Background}

\noindent \textbf{Training Membership Inference.}
Training Membership Inference, also denoted as membership inference attack, aims to predict whether a specific data record has been utilized in the training of a machine learning model. Historically, TMI methods~\cite{shokri2017membership,hui2021practical,leino2020stolen,long2020pragmatic,salem2019ml,truex2019demystifying} have been primarily developed for classification models. The concept is first introduced by Shokri~\cite{shokri2017membership}, who employs a series of shadow models to replicate the behavior of the target model. 

Subsequent research~\cite{chen2020gan,hayes2019logan,hilprecht2019monte,liu2019performing} has expanded these techniques to include generative models, such as GAN and VAE. For instance, LOGAN~\cite{hayes2019logan} operates under the assumption that the GAN's discriminator is more likely to assign a higher probability to training members. Hilprecht~\cite{hilprecht2019monte} employs records within close proximity of a target record to approximate the probability of this record being a member. GAN-Leak~\cite{chen2020gan} posits that members are more closely aligned with the distribution of generated samples.

While these methods have shown promising results in GAN and VAE~\cite{duan2023diffusion}, they have demonstrated suboptimal performance when applied to diffusion models~\cite{duan2023diffusion}. This is primarily due to the challenges associated with the inherent randomness in diffusion models. 
Recently, there have been advancements in Training Membership Inference methods specifically tailored for diffusion models. A notable contribution is from Matsumoto~\cite{matsumoto2023membership}, who introduced a pioneering TMI method for diffusion models. This approach utilizes the training loss of the model and selects a timestep where the divergence of loss between members and nonmembers is greatest to perform the TMI task. SecMI~\cite{duan2023diffusion} proposes the concept of t-error, which represents the posterior estimation error at timestep $t$, assuming that the t-error of members is smaller than that of nonmembers. 
Despite these advancements, these diffusion-specific TMI methods still exhibit certain limitations. First, their performance has not reached a level that is robust enough for practical applications. Second, the label-only predictions generated by these methods lack interpretability, which may lead to a lack of trust from users.

\noindent \textbf{Diffusion Model.}
Denoising Diffusion Probabilistic Models (DDPM)~\cite{ho2020denoising,song2020score} have emerged as a novel branch of generative models. With the impressive ability to approximate the distribution of image data, diffusion models have made breakthroughs in many visual content generation tasks~\cite{rombach2022high,ruiz2023dreambooth,liu2023mfr,wu2023tune,du2023dae,kawar2023imagic,bhunia2023person} and have broken the long-term domination of GANs~\cite{dhariwal2021diffusion,maze2023diffusion,muller2022diffusion}. 
DDPMs consist of a forward and a reverse process. The forward process, also named as the diffusion process, gradually adds Gaussian noise to the input image $x_0$ in $T$ time steps according to a predefined variance schedule $\beta_1, ..., \beta_{T}$:
\begin{equation}
    q(x_t|x_{t-1})=\mathcal{N}(x_t;\sqrt{1-\beta_t}x_{t-1},\beta_t \mathbf{I})
\end{equation}
Let $\alpha_t=1-\beta_t$ and $\bar{\alpha}_t={\textstyle \prod_{s=1}^{t}\alpha_s}$, this process can be simplified to:
\begin{equation}
\label{eq:forward}
    q(x_t|x_0)=\mathcal{N}(x_t;\sqrt{\bar{\alpha}_t}x_0, (1-\bar{\alpha}_t)\mathbf{I})
\end{equation}
When t is large enough, the $\bar{\alpha}$ is approaching 0, making $x_t$ an isotropic Gaussian noise.

The reverse process aims to recover the data distribution from the Gaussian noise. The reverse process in one step can be represented as: 
\begin{equation}
\label{eq:reverse}
    p_{\theta}(x_{t-1}|x_t)=\mathcal{N}(x_{t-1};\mu_{\theta}(x_t,t),\Sigma_{t})
\end{equation}
where $\Sigma_{t}$ is a constant depending on the variance schedule $\beta_t$ and $\mu_{\theta}(x_t,t)$ is determined by a neural network:
\begin{equation}
\mu_{\theta}(x_t, t)=\
    \frac{1}{\sqrt{\alpha_t}}(x_t-\frac{\beta_t}{\sqrt{1-\bar{\alpha}_t}}\epsilon_{\theta}(x_t,t))
\end{equation}
By recursively leveraging the reverse step, Gaussian noise can be recovered to the original image. To train the DDPM, we first sample an image $x_0$, a timestep $t$ and a random noise $\epsilon\sim\mathcal{N}(0, \mathbf{I})$. We can obtain a noisy image $x_t$ using the forward process (Equation \ref{eq:forward}). We then input both the noisy image $x_t$ and the timestep $t$ into a U-Net~\cite{ronneberger2015u} $\epsilon_{\theta}$ to predict the noise within $x_t$. The optimization objective for the denoising U-Net can be written as:
\begin{equation}
\mathcal{L}=\mathbb{E}_{t,x_0,\epsilon}[||\epsilon-\epsilon_\theta(x_t,t)||_2^2]
\end{equation}
To control the randomness of the reverse process, DDIM \cite{song2020denoising} adjusts the noise added in each step:
\begin{subequations}
\label{eq:ddim}
\begin{gather}
        x_{t-1}=\sqrt{\bar{\alpha}_{t-1}}\hat{x}_0+\sqrt{1-\bar{\alpha}_{t-1}}\hat{\epsilon}  \label{Za}\\
        \hat{x}_0=\frac{1}{\sqrt{\bar{\alpha}_t}}(x_t-\sqrt{1-\bar{\alpha}_t}\epsilon_{\theta}(x_t,t))  \label{Zb}\\
        \sqrt{1-\bar{\alpha}_{t-1}}\hat{\epsilon}=\sqrt{1-\bar{\alpha}_{t-1}-\sigma_t^2}\epsilon_{\theta}(x_t,t)+\sigma_t \epsilon \label{Zc} 
\end{gather}
\end{subequations}
By adjusting $\sigma_t$, we can control the random noise added in each reverse step. When $\sigma_t=0$, the reverse process becomes a deterministic process.

In the context of our study, the diffusion model serves as the target model. Considering the practical implications of the TMI task, it is assumed that we only have black-box access to the victim diffusion model, which also follows the setting in~\cite{duan2023diffusion}. This implies that we can only send queries to the victim model and receive corresponding responses while we have no access to the internal weights of the target diffusion model.

\section{Method}
\label{sec:method}
\subsection{Overview}
\begin{figure*}[t]
\centering
\includegraphics[width=0.99\linewidth]{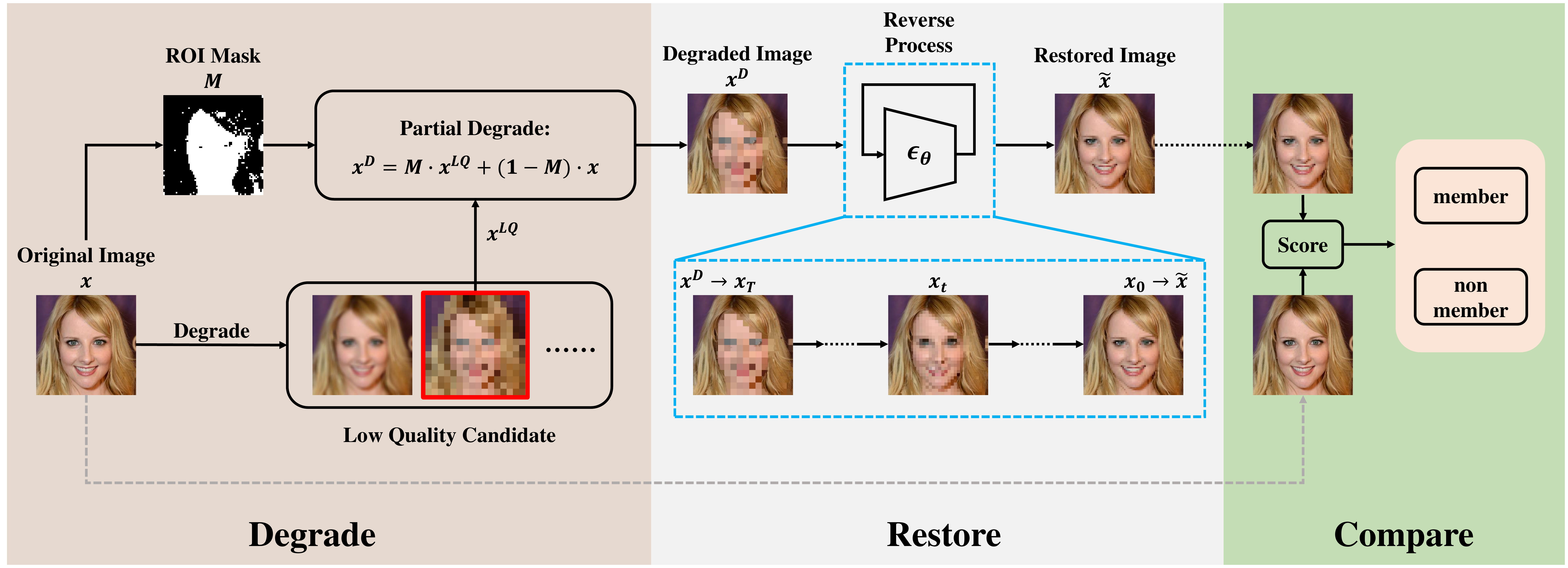}
\caption{The framework of our proposed \textbf{DRC}. Given the original image record $x$, we first obtain a partially degraded image $x^D$ by degrading the face area of the original image $x$. We then exploit the generative priors of the diffusion model $\epsilon_{\theta}$ to restore $x^D$, acquiring a restored image $\Tilde{x}$. Finally, we compare the original image $x$ and its restored image $\Tilde{x}$ and compute a membership score to define whether $x$ is in the training set. 
}
\label{method_framework}
\end{figure*}

Given the original image $x$, the goal of training membership inference is to identify whether $x$ is in the training set of the target diffusion model $\epsilon_{\theta}$. In this work, we propose the \textbf{D}egrade-\textbf{R}estore-\textbf{C}ompare (\textbf{DRC}) framework, which leverages the inherent generative priors in the target diffusion model to infer the training membership with enhanced accuracy and comprehensibility of the results. The architecture of DRC is shown in Figure \ref{method_framework}. 
First, we obtain a partially degraded image $x^D$ by degrading the face area of the original image $x$ (Section \ref{method:degrade}). We then exploit the generative priors of the target model to restore $x^D$, thereby obtaining a restored image $\Tilde{x}$ (Section \ref{method:restore}). 
Image restoration is an ill-posed problem: a single degraded image can correspond to multiple potential restored images. If the restored image $\Tilde{x}$ closely resembles the original image, there is a high probability that the model has encountered $x$ during its training phase. Thus, we adopt a pretrained network to extract the semantic representations to compare the original image $x$ with the restored output $\Tilde{x}$, thereby determining whether $x$ is part of the training set (Section \ref{method:compare}).

\subsection{Degrade}
\label{method:degrade}
As introduced in Figure~\ref{motivation}, training samples manifest as more pronounced peaks compared to unseen samples, which means the degraded training samples can be restored to the original peaks, even with degradation of a high degree. However, it remains too challenging to reproduce the original image $x$ from a fully degraded image. Therefore, we focus on the degradation within the region of interest (ROI), the inner face region in human face images for example. As illustrated in Figure \ref{method_framework}, we first predict the ROI in the original image, represented as the ROI mask $M$. We then apply various types of degradation (e.g., blurring, downsampling) to $x$ to obtain a set of low quality candidates. $x_{LQ}$ is derived by selecting one candidate from this set. (We analyze the effects of different degradation types in Section \ref{exp:ablation}). Finally, we utilize $M$ to integrate the original image $x$ and its corresponding fully degraded $x^{LQ}$. This process can be represented as: 
\begin{equation}
    x_D=M~\cdot x^{LQ} + (1-M)~\cdot x
\end{equation}


Specifically, we use DINO~\cite{caron2021emerging} to predict the ROI mask $M$. DINO is a self-supervised learning method which can extract meaningful visual features and producing attention scores for each pixel. We construct the mask $M$ for the top $p$\% of pixels with highest attention scores. The degree of degradation can be modulated by adjusting the mask ratio. (The impact of varying degrees of degradation is discussed in Section \ref{exp:ablation}).

\begin{algorithm*}[ht!]
    \caption{Restore}
    \label{alg:restore}
\begin{algorithmic}
\STATE \textbf{Input:} degraded image $x^D$, original image $x$, ROI mask $M$, target diffusion model $\epsilon_{\theta}$
\STATE \textbf{Output:} restored image $\Tilde{x}$
\STATE $x_T\leftarrow x^D$
\FOR{$t=T,\dots,1$}
\STATE Sample $\epsilon\sim\mathcal{N}(0, \mathbf{I})$
\STATE $x'\leftarrow \sqrt{\bar{\alpha}_t}x+\sqrt{1-\bar{\alpha}_t}\epsilon$
\STATE $x_t\leftarrow M~\cdot x_t + (1-M)~\cdot x'$
\STATE $\hat{x}_0\leftarrow\frac{1}{\sqrt{\bar{\alpha}_t}}(x_t-\sqrt{1-\bar{\alpha}_t}\epsilon_{\theta}(x_t,t))$
\STATE $x_{t-1}\leftarrow\sqrt{\bar{\alpha}_{t-1}}\hat{x}_0+\sqrt{1-\bar{\alpha}_{t-1}}\epsilon_{\theta}(x_t,t)$
\ENDFOR
\STATE $\Tilde{x}\leftarrow x_0$
\STATE \textbf{return} $\Tilde{x}$
\end{algorithmic}
\end{algorithm*}

\subsection{Restore}
\label{method:restore}
We leverage the generative priors of the target diffusion model $\epsilon_{\theta}$ to restore the degraded image $x^D$. More precisely, our aim is to restore the masked ROI region, conditioned on the surrounding unmasked pixels. A common approach is to treat the degraded image $x^D$ as $x_T$, which serves as the initial point of the reverse process. However, this approach results in the output irrelevant to the conditional unmasked pixels. This discrepancy arises because the reverse process comprises $T$ denoising steps and the model progressively forgets the initial conditions with the step increasing. To overcome this problem, in each denoising step, we replace the surrounding pixels with $x'$ (the prior of the original image $x$) to force the denoising output aligned with the condition. 
This process can be represented as:
\begin{equation}
\label{eq:replace}
    x_t\leftarrow M~\cdot x_t + (1-M)~\cdot x’
\end{equation}
Due to the stochastic nature of the diffusion model, we may produce dissimilar restorations for member data, which will make member data and non-member data less separable.
To eliminate the randomness of the diffusion model,  we use deterministic DDIM sampling when $\sigma_t=0$ (Equation \ref{eq:ddim}). Our restoration process is also summarized in Algorithm \ref{alg:restore}.

\subsection{Comparison}
\label{method:compare}
By comparing the restored image $\Tilde{x}$ with the original image $x$, a membership score is assigned to $x$. The more similar these two images are, the larger the membership score is, which demonstrates a higher probability that $x$ is a member in the training set. The memory mechanism of diffusion models is predicted on semantic associations rather than pixel-level details (which will be demonstrated in Section \ref{exp:further}). Therefore, we compare these two images from a semantic perspective. Specifically, we first utilize a pretrained semantic encoder to obtain the semantic representations of these two images. We then compute the cosine similarity of the semantic representations as the membership score. This process can be represented as:
\begin{equation}
    score=\frac{SE(x)\cdot SE(\Tilde{x})}{||SE(x)||||SE(\Tilde{x})||}
\end{equation}
where $SE(\cdot)$ is the semantic encoder. In this paper, we adopt a face recognition network~\cite{deng2019arcface} as the semantic encoder for human face images. We adopt CLIP's image encoder~\cite{radford2021learning} as the semantic encoder for natural images.

\section{Experiment}
\subsection{Experimental Setup}
\noindent \textbf{Datasets and Diffusion Models.} 
For natural image generation, we evaluate two commonly used datasets: Cifar10 and Cifar100~\cite{krizhevsky2009learning} datasets. We train the diffusion model on Cifar10 and Cifar100 following the setting of~\cite{duan2023diffusion}. For face generation, we evaluate the CelebA~\cite{karras2018progressive} and Flickr-Faces-HQ~(FFHQ)~\cite{karras2019style} datasets in our experiments. We train the diffusion model on the CelebA~\cite{karras2018progressive} dataset following the architecture of~\cite{rombach2022high}. We also adopt a widely-used open-sourced diffusion model~\footnote{\href{https://github.com/CompVis/latent-diffusion}{https://github.com/CompVis/latent-diffusion}} trained on the FFHQ dataset. Note that this model is officially provided. For Cifar10, Cifar100 and CelebA datasets, we randomly select half of the training sets to train the diffusion models, tagged as \textbf{members}. The remaining half is tagged as \textbf{non-members}. For the open-sourced diffusion model in FFHQ, the member set is the whole FFHQ dataset~\footnote{It should be noted that the diffusion model in FFHQ may not be trained in the whole FFHQ dataset. Some images are used for validation only. However, these validation images are not reported. Therefore, we label the whole FFHQ as the member set}. We choose the CelebA dataset as the non-member set. The experimental results reported in the following are all based on this label setting.

\noindent \textbf{Evaluation Metrics.} 
To evaluate the performance of our proposed method, we choose widely used metrics~\cite{hu2022membership} including Accuracy~(Acc), Precision, Recall, and Area-Under-the-ROC-curve~(AUC). Acc is the ratio of correctly predicted samples to the total samples. Precision is the fraction of samples classified as members that are indeed members of the training set. Recall is the fraction of the members that are correctly classified as members. For the calculation of the AUC metric, we refer to~\cite{swets1988measuring} which also gives a detailed explanation about AUC. We also consider the True Positive Rate (TPR) at extremely low False Positive Rate (FPR) to gauge the confidence level of the prediction, as suggested by Carlini et al.~\cite{carlini2022membership}
TPR@1\%FPR and TPR@0.1\%FPR denote the True Positive Rate when the False Positive Rate is set at 1\% and 0.1\% respectively.

\begin{table*}[ht]
\centering
\caption{Performance of our proposed method in CelebA and FFHQ. $\uparrow$ represents that the higher the metric, the better the performance. Bold denotes the best result for each metric.}
\begin{tabular}{ccccccccc}
\hline
           & \multicolumn{4}{c}{CelebA} & \multicolumn{4}{c}{FFHQ}      \\
\cmidrule(lr){2-5}\cmidrule(lr){6-9}
 Method   & Acc$\uparrow$   & Recall$\uparrow$ & Precision$\uparrow$ & AUC$\uparrow$   & Acc$\uparrow$   & Recall$\uparrow$ & Precision$\uparrow$ & AUC$\uparrow$   \\ \hline
$\text{GAN-Leak}_\textit{ori}$~\cite{chen2020gan} & 0.511 & 0.048  & 0.632     & 0.491 & 0.511 & 0.054  & 0.628     & 0.496 \\
$\text{GAN-Leak}_\textit{inv}$~\cite{chen2020gan} & 0.645 & 0.634  & 0.648     & 0.669 & 0.544 & \textbf{0.738}  & 0.532     & 0.538 \\
NaiveLoss~\cite{matsumoto2023membership}   & 0.773 & 0.728  & 0.800     & 0.846 & 0.658 & 0.656  & 0.659     & 0.695 \\
SecMI~\cite{duan2023diffusion}     & 0.779 & 0.804  & 0.766     & 0.848 & 0.629 & 0.630  & 0.629     & 0.656 \\ \hline
Ours   & \textbf{0.962} & \textbf{0.962}  & \textbf{0.962}     & \textbf{0.989} & \textbf{0.753} & 0.622  & \textbf{0.843}     & \textbf{0.811} \\ \hline
\end{tabular}
\label{tab:main}
\end{table*}

\begin{table*}[ht]
\centering
\caption{Performance of our proposed method in Cifar10 and Cifar100. $\uparrow$ represents that the higher the metric, the better the performance. Bold denotes the best result for each metric.}
\begin{tabular}{ccccccccc}
\hline
& \multicolumn{4}{c}{Cifar10} & \multicolumn{4}{c}{Cifar100}      \\
\cmidrule(lr){2-5}\cmidrule(lr){6-9}
 Method   & Acc$\uparrow$   & Recall$\uparrow$ & Precision$\uparrow$ & AUC$\uparrow$   & Acc$\uparrow$   & Recall$\uparrow$ & Precision$\uparrow$ & AUC$\uparrow$   \\ \hline
$\text{GAN-Leak}_\textit{ori}$~\cite{chen2020gan} & 0.560 & 0.631 & 0.553 & 0.580 & 0.549 & 0.615 & 0.543 & 0.567 \\
$\text{GAN-Leak}_\textit{inv}$~\cite{chen2020gan} & 0.556 & 0.654 & 0.547 & 0.573 & 0.558 & 0.707 & 0.544 & 0.581 \\
NaiveLoss~\cite{matsumoto2023membership} & 0.761 & 0.821 & 0.732 & 0.822 & 0.732 & 0.816 & 0.698 & 0.797 \\
SecMI~\cite{duan2023diffusion}     & 0.825 & 0.877 & 0.794 & 0.886 & 0.800 & 0.866 & 0.765 & 0.869 \\ \hline
Ours      & \textbf{0.874} & \textbf{0.920} & \textbf{0.843} & \textbf{0.931} & \textbf{0.852} & \textbf{0.916} & \textbf{0.812} & \textbf{0.919} \\ \hline
\end{tabular}
\label{tab:main_cifar}
\vspace{-0.2cm}
\end{table*}

\noindent \textbf{Implementation Details.}
Our DRC has two main hyper-parameters: the mask ratio and the degradation type. The former controls the degraded pixel numbers. The latter defines the corresponding restoration task (such as denoising, inpainting, etc.). 
Due to the varying semantic information contained in natural images versus facial images, we employ distinct mask ratios.
For CelebA and FFHQ datasets, we mask 20\% most important pixels in the original image (mask ratio=0.2). For Cifar10/100, we mask 50\% most important pixels in the original image (mask ratio=0.5). We choose denoising as the restoration task. In these settings, our method achieves the best performance. We also conduct detailed ablations in Section \ref{exp:ablation}.
Besides, we adopt DDIM sampling ($\sigma_t=0$) with an interval of 5 to accelerate the reverse process of diffusion models.
To calculate the Accuracy~(Acc), a threshold is needed. We follow the method in~\cite{duan2023diffusion} to obtain the threshold.

\subsection{Comparison to Baselines}
We compare our method with the current state-of-the-art MIA method, SecMI~\cite{duan2023diffusion}. We also compare with other baseline methods including $\text{GAN-Leak}$~\cite{chen2020gan} and NaiveLoss~\cite{matsumoto2023membership}. As GAN-Leak is designed for GANs and can not be directly used for diffusion models, we follow the adjustments in~\cite{duan2023diffusion} to make it available for diffusion models. (The details can be found in Supplementary Materials.)

\begin{figure*}[t]
\centering
\includegraphics[width=0.99\linewidth]{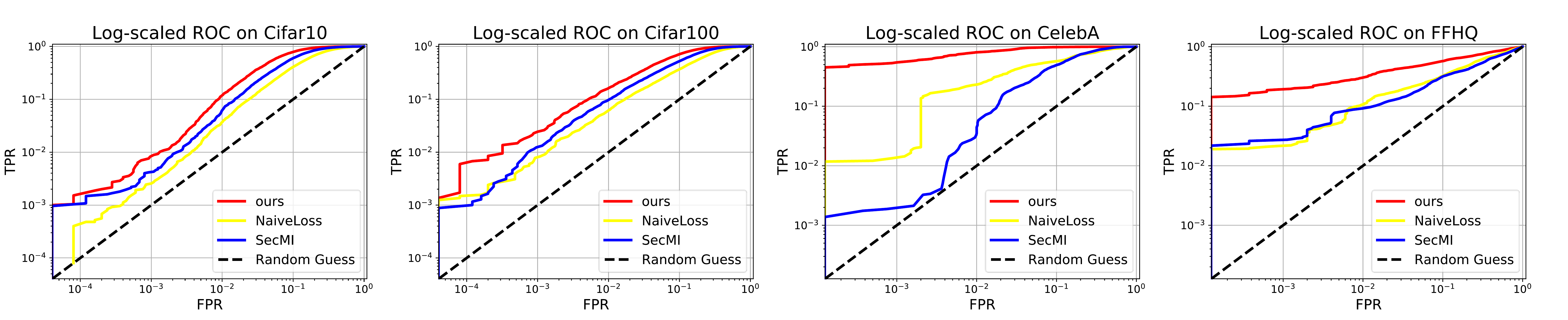}
\caption{The log scaled ROC curves on Cifar10, Cifar100, CelebA and FFHQ datasets. The log scaled ROC provides compelling evidence that our method can substantially perform training membership inference with high prediction confidence. 
}
\label{fig:log_roc}
\end{figure*}

\begin{table*}[t]
\centering
\caption{The True Positive Rate (TPR) at very low False Positive Rate (FPR) of our proposed method. TPR@1\% and TPR@0.1\% stands for TPR when the FPR is set at 1\% and 0.1\% separately. $\uparrow$ represents that the higher the metric, the better the performance. Bold denotes the best result for each metric.}
\begin{tabular}{ccccccccc}
\hline
 & \multicolumn{4}{c}{TPR@1\%FPR(\%)$\uparrow$} & \multicolumn{4}{c}{TPR@0.1\%FPR(\%)$\uparrow$} \\ \cmidrule(lr){2-5}\cmidrule(lr){6-9}
Method& Cifar10 & Cifar100 & CelebA & FFHQ  & Cifar10  & Cifar100 & CelebA & FFHQ  \\ \hline
$\text{GAN-Leak}_\textit{ori}$~\cite{chen2020gan} & 1.28 & 1.97 & 2.22 & 1.80 & 0.12 & 0.18 & 0.51 & 0.55  \\
$\text{GAN-Leak}_\textit{inv}$~\cite{chen2020gan} & 1.31 & 1.89 & 2.50 & 0.63 & 0.09 & 0.21 & 0.22 & 0.01  \\
NaiveLoss~\cite{matsumoto2023membership}          & 3.77 & 6.13 & 23.16 & 11.20 & 0.26 & 0.83 & 1.34 & 2.31  \\
SecMI~\cite{duan2023diffusion}                    & 6.09 & 9.68 & 4.43 & 9.46 & 0.43 & 1.23 & 0.20 & 2.66  \\ \hline
Ours & \textbf{12.20} & \textbf{16.56} & \textbf{80.46}& \textbf{30.85} & \textbf{0.82} & \textbf{2.44} & \textbf{54.52}& \textbf{18.60} \\ \hline
\end{tabular}
\label{tab:main_tprfpr}
\end{table*}

Table \ref{tab:main} and Table \ref{tab:main_cifar} presents the performance of our proposed method and other compared TMI methods. We observe that our method achieves the best performance on the accuracy, precision, and AUC metrics and outperforms baselines by a large margin, which shows the superiority of our proposed methods. We also find that our method is not so effective in the recall metric in FFHQ. This may be attributed to the varying complexity of different images. Though we make our best efforts to control the difficulty of image restoration (by masking areas of the same ratio), the semantic information of masked areas still varies between different images. More precisely, under the condition of masking the same number of pixels, some images have simpler semantic information, while others are too complex for the diffusion model to fully reproduce. This will make the member samples misclassified to nonmember samples, resulting in a decrease in recall metric. We also find that in FFHQ, our method enjoys a high precision of up to 0.843. This is because our method is based on reproduction. Once a sample is fully reproduced, there is a high probability that the model has `seen' the sample during its training stage. We further note that privacy violation detection mandates user trust as a prerequisite, thus the precision metric assumes a more vital role in this context. Besides, we will show (in Section~\ref{exp:ablation}) that by simply aggregating the restoration results of different degradation types, our method can obtain a boost in recall and reach a balance between precision and recall. 
Table \ref{tab:main_tprfpr} presents the True Positive Rate (TPR) at very low False Positive Rate (FPR) of our method. It can be observed that our method demonstrates significant success in both of the two evaluations. We also present the log-scaled ROC curves in Figure~\ref{fig:log_roc}.

\subsection{Ablation Study}
\label{exp:ablation}
\noindent \textbf{Restoration Tasks.} 
Different degradation methods will manipulate the original image to different positions on the learned manifold, which also influences the initial point of the restoration task. The distance between the degraded image and the original image determines the difficulty of the restoration task and thus significantly affects the restoration results. Therefore, we explored various degradations to construct different restoration tasks (including deblur, derain, inpaint, super resolution, and denoise), and the results in CelebA are shown in Table \ref{tab:ablation_degrade_type}. It can be seen that the performance gaps among different restoration tasks are rather small, with a deviation less than 0.01. The denoising method performs the best, while the inpainting method the worst. This is because the training objective of the diffusion model is also to denoise the input image with varying noise levels. The consistency in objective may recall the model's memory of the training sample. Compared to denoising, the inpainting process needs to reproduce the image from an area without signals, which contradicts the training object most and results in suboptimal performance.  

\begin{minipage}{\linewidth}
\vspace{+0.2cm}
\begin{minipage}[t]{0.45\linewidth}
     \makeatletter\def\@captype{table}\makeatother\caption{The result of performing different restoration tasks in CelebA. We have tried five degradation types including deblur, derain, inpaint, super resolution, and denoise.}
\label{tab:ablation_degrade_type}
\resizebox{0.99\linewidth}{!}{
\begin{tabular}{ccccc}
\hline
\begin{tabular}[c]{@{}c@{}}Restoration\\ Task\end{tabular}                 & Acc$\uparrow$   & Recall$\uparrow$ & Precision$\uparrow$ & AUC$\uparrow$   \\ \hline
deblur           & 0.959 & 0.958  & 0.960     & 0.988 \\
derain           & 0.960 & \textbf{0.964}  & 0.956     & \textbf{0.989} \\
inpaint          & 0.954 & 0.954  & 0.954     & 0.988 \\
super resolution & 0.959 & 0.960  & 0.958     & \textbf{0.989} \\ 
denoise          & \textbf{0.962} & 0.962  & \textbf{0.962}     & \textbf{0.989} \\ \hline
\end{tabular}
}
\end{minipage}
  \quad
  \begin{minipage}[t]{0.45\linewidth}
        \makeatletter\def\@captype{table}\makeatother\caption{The results of simple aggregation methods FFHQ. We carry the aggregation by computing the mean and median of the results of different restoration tasks.}
\label{tab:ablation_aggregation}
\resizebox{0.99\linewidth}{!}{
\begin{tabular}{ccccc}
\hline
\begin{tabular}[c]{@{}c@{}}Aggregation\\ Strategy\end{tabular} & Acc$\uparrow$   & Recall$\uparrow$ & Precision$\uparrow$ & AUC$\uparrow$   \\ \hline
mean                                                      & \textbf{0.774} & \textbf{0.768} & 0.777 & \textbf{0.842} \\
median                                                         & 0.759 & 0.738 & 0.770 & 0.823 \\
denoise only                                                   & 0.753 & 0.622 & \textbf{0.843} & 0.811 \\ \hline
\end{tabular}
}
   \end{minipage}
\vspace{+0.3cm}
\end{minipage}

\begin{figure*}[b]
	\begin{center}
		\centerline{\includegraphics[width=0.80\textwidth]{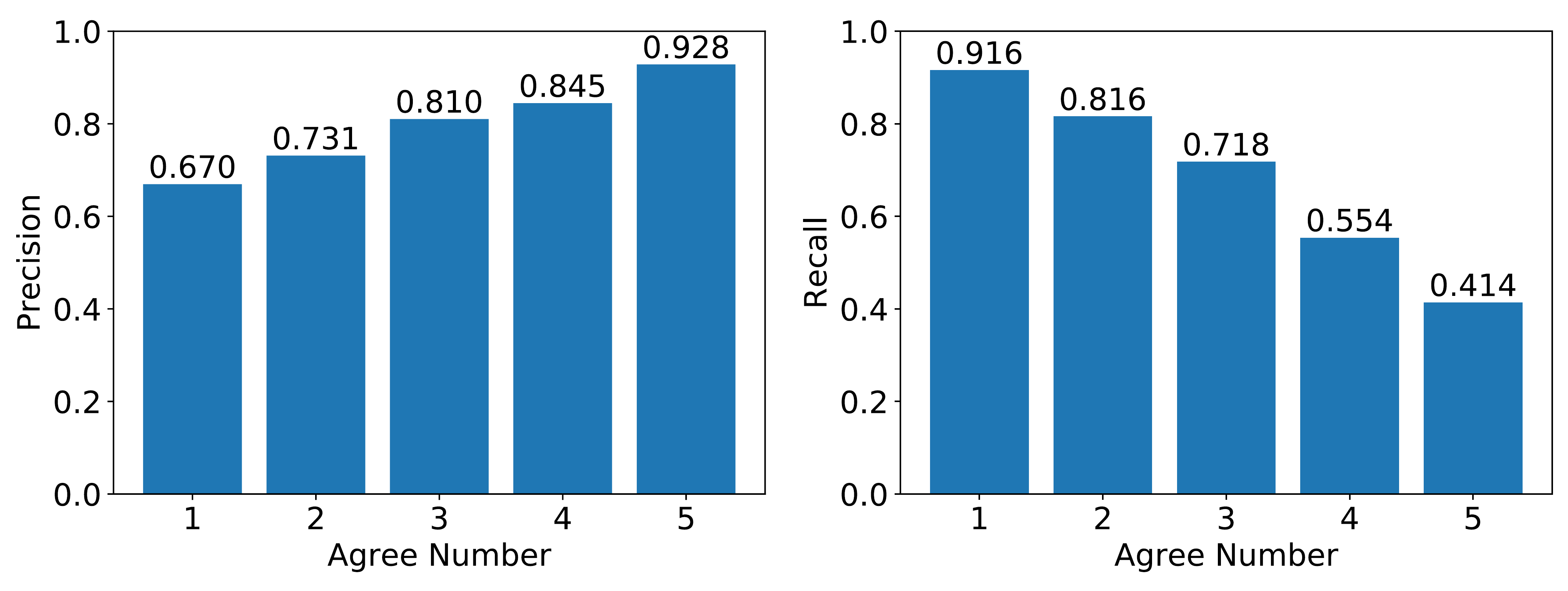}}
		\caption{We design an aggregation based on the agree numbers of different restoration tasks. We plot the Precision and Recall in different agree numbers in FFHQ.
        }
	\label{fig:ablation_AP_AR}
	\end{center}
\end{figure*}

\noindent \textbf{Aggregation of Restoration Methods.}
We attempt to further aggregate the results of different restoration methods. We first explore aggregating these results by simply computing the mean/median of the five restoration methods' member scores. The results are shown in Table \ref{tab:ablation_aggregation}. We observe that in FFHQ, the aggregation methods (both mean and median) can balance the recall and precision metrics. Considering the high precision requirements in practical privacy violation detection, we design an aggregation method based on voting. Specifically, we consider a sample as member only if a certain number of the restoration methods classify this sample as member. We refer to this number as the `agree number'. Figure \ref{fig:ablation_AP_AR} demonstrates the changes in precision with different agree numbers. It can be observed that as the agree number increases, the precision boosts from 0.670 to 0.928, which demonstrates promising application potential in privacy violation detection. We believe that if the agree number continues to increase, we will obtain a higher precision. We also plot changes in recall with different agree numbers (also in Figure \ref{fig:ablation_AP_AR}). We can observe that the recall metric achieves its peak value when there is only one restoration method that classifies the sample as member. This is because with the agree number increasing, the criterion for a sample to be classified as member becomes more and more strict. For example, when agree number reaches 5, a sample will be labeled as member only if all the five restoration methods label it as member. By adjusting the agree number, we can control the tendency towards precision and recall, allowing for more flexible usage in real world applications.  

\begin{figure*}[t]
	\begin{center}
		\centerline{\includegraphics[width=0.98\textwidth]{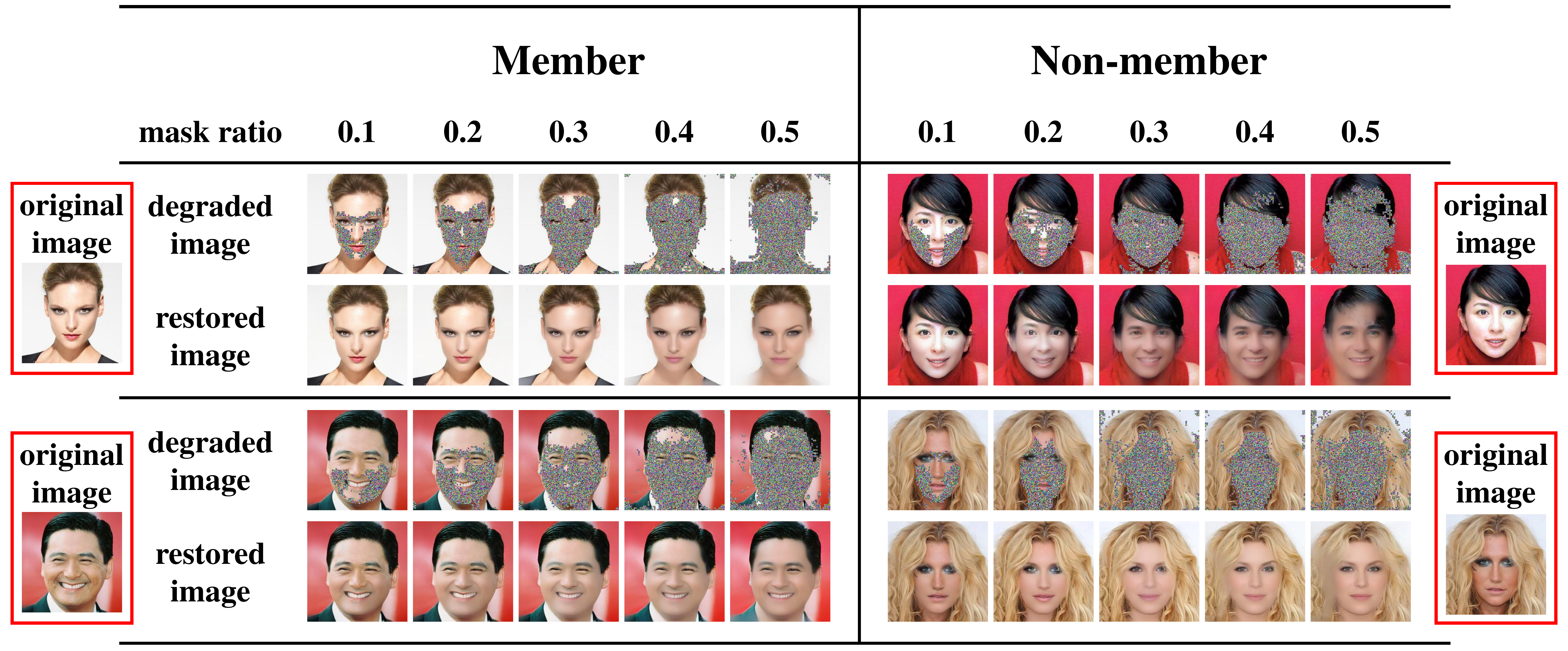}}
		\caption{The degraded and restored images of our method in different mask ratios. We show two members and non-members with the mask ratio varying from 0.1 to 0.5. More results can be found in supplementary materials.
        }
	\label{fig:ablation_mask_ratio}
	\end{center}
\vspace{-1.0cm}
\end{figure*}

\begin{wraptable}{r}{0.5\linewidth}
\vskip -0.47 in
\caption{Ablations on mask ratio in CelebA. We evaluate the performance with the mask ratio from 0.1 to 0.5.}
\centering
\resizebox{0.98\linewidth}{!}{
\begin{tabular}{ccccc}
\hline
mask ratio     & Acc$\uparrow$   & Recall$\uparrow$ & Precision$\uparrow$ & AUC$\uparrow$   \\ \hline
        0.1 & 0.541 & 0.492  & 0.545     & 0.548 \\
        0.2 & \textbf{0.962} & \textbf{0.962}  & \textbf{0.962}     & \textbf{0.989} \\
        0.3 & 0.945 & 0.938  & 0.951     & 0.980 \\
        0.4 & 0.889 & 0.832  & 0.939     & 0.936 \\
        0.5 & 0.816 & 0.704  & 0.907     & 0.869 \\ \hline
\end{tabular}
}
\vskip -0.3 in
\label{tab:ablation_mask_ratio}
\end{wraptable}

\noindent \textbf{Degradation Degree.} 
The mask ratio is used to control the image degradation degree and thereby adjust the difficulty of image restoration. We investigate the performance under different mask ratios. The qualitative and quantitative results are shown in Figure \ref{fig:ablation_mask_ratio} and Table \ref{tab:ablation_mask_ratio}, respectively. It can be observed that as the mask ratio increases, the performance of our proposed method first increases (reaches its optimal level at 0.2) and then decreases. 
The obtained experimental results are consistent with the fundamental intuition outlined in our introduction, which posits that compared to unseen samples, training samples can be restored to their original state with a higher degree of degradation. The mask ratio is indicative of the degradation degree. At lower degradation degrees, both degraded training and unseen samples can be restored. As the degradation level increases within a certain range, the model can only restore training samples, resulting in optimal discriminative performance. However, when the degradation level is beyond the generative priors, the model is unable to restore either type of degraded sample, leading to a decline in discriminative capability.

Therefore, we believe there exists a point where the masked semantic area and unmasked semantic priors reach a balance, achieving the optimal performance of our proposed method. 

\begin{figure*}[t]
	\begin{center}
		\centerline{\includegraphics[width=0.98\textwidth]{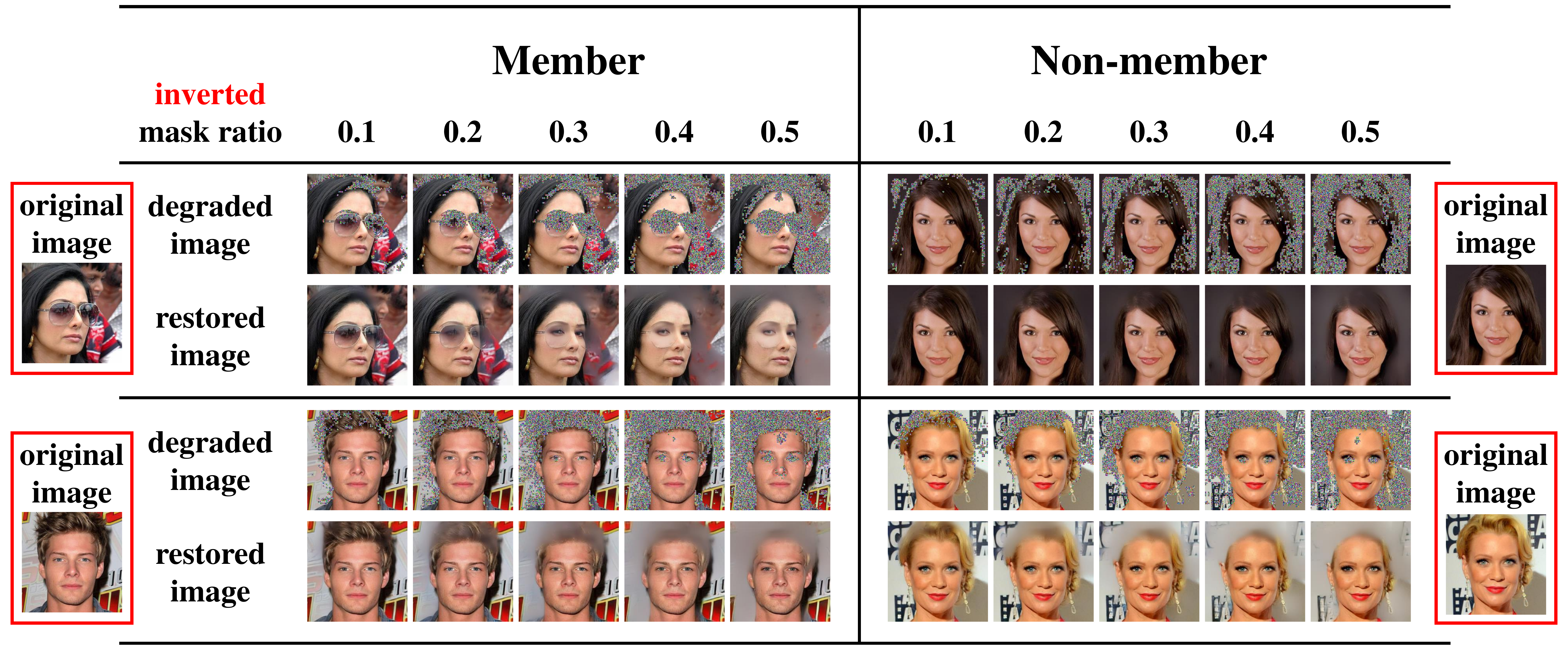}}
		\caption{The degraded and restored images of our method in different \textbf{inverted} mask ratios. We show two members and non-members with the mask ratio varying from 0.1 to 0.5. More results can be found in supplementary materials.
        }
	\label{fig:ablation_inverted_mask_ratio}
	\end{center}
\vspace{-0.5cm}
\end{figure*}

\begin{table*}[t]
\caption{Quantitative results with \textbf{inverted} mask ratios varying from 0.2 to 0.5. In addition to the identity metric (CSIM), we also employ pixel level metrics ($\text{L}_1$, MSE) to distinguish between members and non-members.}
\centering
\resizebox{\textwidth}{!}{
\begin{tabular}{ccccccccccccc}
\hline
\multirow{2}{*}{\begin{tabular}[c]{@{}c@{}}inverted\\ mask ratio\end{tabular}}       & \multicolumn{4}{c}{CSIM}          & \multicolumn{4}{c}{$\text{L}_1$}            & \multicolumn{4}{c}{MSE}            \\
\cmidrule(lr){2-5}\cmidrule(lr){6-9}\cmidrule(lr){10-13}
& Acc$\uparrow$   & Recall$\uparrow$ & Precision$\uparrow$ & AUC$\uparrow$   & Acc$\uparrow$   & Recall$\uparrow$ & Precision$\uparrow$ & AUC$\uparrow$   & Acc$\uparrow$   & Recall$\uparrow$ & Precision$\uparrow$ & AUC$\uparrow$   \\ \hline
0.2 & 0.585 & 0.844  & 0.556     & 0.608 & 0.655 & \textbf{0.778}  & 0.624     & \textbf{0.699} & \textbf{0.662} & 0.648  & \textbf{0.667}     & \textbf{0.702} \\
0.3 & 0.606 & \textbf{0.852}  & 0.571     & 0.638 & \textbf{0.656} & 0.658  & 0.655     & 0.694 & 0.645 & \textbf{0.792}  & 0.612     & 0.691 \\
0.4 & \textbf{0.637} & 0.736  & 0.614     & 0.679 & 0.643 & 0.630  & 0.647     & 0.679 & 0.637 & 0.594  & 0.650     & 0.672 \\
0.5 & 0.540 & 0.746  & \textbf{0.626}     & \textbf{0.698} & 0.629 & 0.522  & \textbf{0.664}     & 0.657 & 0.619 & 0.582  & 0.629     & 0.649 \\ \hline
\end{tabular}
}
\label{tab:ablation_inverse_mask}
\vspace{-0.3cm}
\end{table*}

\subsection{Further Analysis}
\label{exp:further}
\noindent \textbf{Inverted Mask areas.} 
Facial privacy is closely intertwined with our personal privacy~\cite{van2020ethical,he2023diff}. Consequently, we express particular interest in model responses to human faces. We further explore what diffusion models tend to memorize in human face images.
In our method, we mask pixels that are most \textbf{important} in face images. In this experiment, we invert the mask area, i.e. we mask pixels that are most \textbf{unimportant} in face images. With the inverted mask ratio increasing, we first mask the background and then remove some of the hair. It should be noted that in this experiment, the human face of each image is almost left unmasked.  
If the background of the restored image is similar to the original image, it indicates that the model memorizes every pixel in the image without discrimination, including both the background and the face geometry.
On the contrary, if diffusion models can not restore the masked background, it indicates that the model focuses more on the face area. This tendency to memorize face identity reveals an even higher risk of the privacy leakage.
The quantitative and qualitative results are shown in Figure \ref{fig:ablation_inverted_mask_ratio} and Table \ref{tab:ablation_inverse_mask}.
In our method, we use similarity in identity to distinguish the members and non-members (termed as CSIM).
However, this metric may not work as the identity information in degraded images is not ruined in this experiment. Thus, we also use pixel level loss ($\text{L}_1$ loss and MSE loss) to discriminate between members and non-members.
From Table \ref{tab:ablation_inverse_mask}, we can observe that pixel level metrics perform slightly better than identity metric (CSIM) . However, both these metrics show unsatisfactory performance with the best AUC achieving only 0.702. This indicates that diffusion models are inclined to ignore information unrelated to identity, such as background, and decorations. Figure \ref{fig:ablation_inverted_mask_ratio} further verifies this indication. In the upper left of Figure \ref{fig:ablation_inverted_mask_ratio}, the sunglasses are restored to eyes while the textures in the background are ignored by the diffusion models, leaving only a blurry area. 


\vspace{-0.2cm}
\section{Conclusion}
In this study, we introduce DRC, an innovative training membership inference framework for diffusion models by exploiting the intrinsic reasoning and generative priors of the diffusion model. 
Specifically, we first perform degradation in the high attention area of the original image. Then, we exploit the generative priors to restore the degraded area. Finally, we compare the original image with its restored counterpart to determine its membership. 
Our experiment results validate the proficiency of our methodology in accurately distinguishing between members and non-members. Moreover, the underlying mechanism of our approach is intuitively comprehensible, which enhances the credibility of our predicted outcomes for users. As a user-friendly technique, we sincerely anticipate that our approach will make a substantial contribution to the safeguarding of individual privacy in real-world applications. 

\vspace{-0.2cm}
\section*{Limitation and Ethical Statement}
In this paper, we develop a novel TMI framework for diffusion models. While our approach demonstrates effectiveness, it is hard to establish sufficient theoretical evidence. In future work, we will try to provide a rigorous mathematical proof to validate the theoretical foundation.
The goal of this research is to develop a technique to identify whether a specific sample is part of the training dataset. The method we propose enjoys numerous advantageous applications, including privacy violation detection and model privacy assessment. Despite the potential for malicious entities to exploit our method for privacy attacks, we emphasize that such attacks can be mitigated by privacy protection techniques, such as differential privacy. Moreover, we intend to release our code to facilitate the advancement of privacy protection technology. 

%
%
\bibliographystyle{splncs04}
\bibliography{main}

\begin{thebibliography}{10}
\providecommand{\url}[1]{\texttt{#1}}
\providecommand{\urlprefix}{URL }
\providecommand{\doi}[1]{https://doi.org/#1}

\bibitem{bhunia2023person}
Bhunia, A.K., Khan, S., Cholakkal, H., Anwer, R.M., Laaksonen, J., Shah, M., Khan, F.S.: Person image synthesis via denoising diffusion model. In: Proceedings of the IEEE/CVF Conference on Computer Vision and Pattern Recognition. pp. 5968--5976 (2023)

\bibitem{brittain2023getty}
Brittain, B.: Getty images lawsuit says stability ai misused photos to train ai. Reuters, Feb 6th  (2023)

\bibitem{carlini2022membership}
Carlini, N., Chien, S., Nasr, M., Song, S., Terzis, A., Tramer, F.: Membership inference attacks from first principles. In: 2022 IEEE Symposium on Security and Privacy (SP). pp. 1897--1914. IEEE (2022)

\bibitem{caron2021emerging}
Caron, M., Touvron, H., Misra, I., J{\'e}gou, H., Mairal, J., Bojanowski, P., Joulin, A.: Emerging properties in self-supervised vision transformers. In: Proceedings of the IEEE/CVF international conference on computer vision. pp. 9650--9660 (2021)

\bibitem{chen2020gan}
Chen, D., Yu, N., Zhang, Y., Fritz, M.: Gan-leaks: A taxonomy of membership inference attacks against generative models. In: Proceedings of the 2020 ACM SIGSAC conference on computer and communications security. pp. 343--362 (2020)

\bibitem{deng2019arcface}
Deng, J., Guo, J., Xue, N., Zafeiriou, S.: Arcface: Additive angular margin loss for deep face recognition. In: Proceedings of the IEEE/CVF conference on computer vision and pattern recognition. pp. 4690--4699 (2019)

\bibitem{dhariwal2021diffusion}
Dhariwal, P., Nichol, A.: Diffusion models beat gans on image synthesis. Advances in neural information processing systems  \textbf{34},  8780--8794 (2021)

\bibitem{du2023dae}
Du, C., Chen, Q., He, T., Tan, X., Chen, X., Yu, K., Zhao, S., Bian, J.: Dae-talker: High fidelity speech-driven talking face generation with diffusion autoencoder. In: Proceedings of the 31st ACM International Conference on Multimedia. pp. 4281--4289 (2023)

\bibitem{duan2023diffusion}
Duan, J., Kong, F., Wang, S., Shi, X., Xu, K.: Are diffusion models vulnerable to membership inference attacks? International Conference on Machine Learning  (2023)

\bibitem{edwards2022artists}
Edwards, B.: Artists stage mass protest against ai-generated artwork on artstation. Ars Technica, December  (2022)

\bibitem{goodfellow2014generative}
Goodfellow, I., Pouget-Abadie, J., Mirza, M., Xu, B., Warde-Farley, D., Ozair, S., Courville, A., Bengio, Y.: Generative adversarial nets. Advances in neural information processing systems  \textbf{27} (2014)

\bibitem{hayes2019logan}
Hayes, J., Melis, L., Danezis, G., De~Cristofaro, E.: Logan: Membership inference attacks against generative models. In: Proceedings on Privacy Enhancing Technologies (PoPETs). vol.~2019, pp. 133--152. De Gruyter (2019)

\bibitem{he2023diff}
He, X., Zhu, M., Chen, D., Wang, N., Gao, X.: Diff-privacy: Diffusion-based face privacy protection. arXiv preprint arXiv:2309.05330  (2023)

\bibitem{hilprecht2019monte}
Hilprecht, B., H{\"a}rterich, M., Bernau, D.: Monte carlo and reconstruction membership inference attacks against generative models. Proc. Priv. Enhancing Technol.  \textbf{2019}(4),  232--249 (2019)

\bibitem{ho2020denoising}
Ho, J., Jain, A., Abbeel, P.: Denoising diffusion probabilistic models. Advances in neural information processing systems  \textbf{33},  6840--6851 (2020)

\bibitem{hu2022membership}
Hu, H., Salcic, Z., Sun, L., Dobbie, G., Yu, P.S., Zhang, X.: Membership inference attacks on machine learning: A survey. ACM Computing Surveys (CSUR)  \textbf{54}(11s),  1--37 (2022)

\bibitem{hui2021practical}
Hui, B., Yang, Y., Yuan, H., Burlina, P., Gong, N.Z., Cao, Y.: Practical blind membership inference attack via differential comparisons. In: ISOC Network and Distributed System Security Symposium (NDSS) (2021)

\bibitem{karras2018progressive}
Karras, T., Aila, T., Laine, S., Lehtinen, J.: Progressive growing of gans for improved quality, stability, and variation. In: International Conference on Learning Representations (2018)

\bibitem{karras2019style}
Karras, T., Laine, S., Aila, T.: A style-based generator architecture for generative adversarial networks. In: Proceedings of the IEEE/CVF conference on computer vision and pattern recognition. pp. 4401--4410 (2019)

\bibitem{kawar2023imagic}
Kawar, B., Zada, S., Lang, O., Tov, O., Chang, H., Dekel, T., Mosseri, I., Irani, M.: Imagic: Text-based real image editing with diffusion models. In: Proceedings of the IEEE/CVF Conference on Computer Vision and Pattern Recognition. pp. 6007--6017 (2023)

\bibitem{kingmaauto}
Kingma, D.P., Welling, M.: Auto-encoding variational $\{$Bayes$\}$. In: Int. Conf. on Learning Representations

\bibitem{krizhevsky2009learning}
Krizhevsky, A., Hinton, G., et~al.: Learning multiple layers of features from tiny images  (2009)

\bibitem{lee2023stable}
LEE, T.: Stable diffusion copyright lawsuits could be a legal earthquake for ai— ars technica (2023)

\bibitem{leino2020stolen}
Leino, K., Fredrikson, M.: Stolen memories: Leveraging model memorization for calibrated $\{$White-Box$\}$ membership inference. In: 29th USENIX security symposium (USENIX Security 20). pp. 1605--1622 (2020)

\bibitem{liu2023mfr}
Liu, J., Wang, X., Fu, X., Chai, Y., Yu, C., Dai, J., Han, J.: Mfr-net: Multi-faceted responsive listening head generation via denoising diffusion model. In: Proceedings of the 31st ACM International Conference on Multimedia. pp. 6734--6743 (2023)

\bibitem{liu2019performing}
Liu, K.S., Xiao, C., Li, B., Gao, J.: Performing co-membership attacks against deep generative models. In: 2019 IEEE International Conference on Data Mining (ICDM). pp. 459--467. IEEE (2019)

\bibitem{long2020pragmatic}
Long, Y., Wang, L., Bu, D., Bindschaedler, V., Wang, X., Tang, H., Gunter, C.A., Chen, K.: A pragmatic approach to membership inferences on machine learning models. In: 2020 IEEE European Symposium on Security and Privacy (EuroS\&P). pp. 521--534. IEEE (2020)

\bibitem{matsumoto2023membership}
Matsumoto, T., Miura, T., Yanai, N.: Membership inference attacks against diffusion models. arXiv preprint arXiv:2302.03262  (2023)

\bibitem{maze2023diffusion}
Maz{\'e}, F., Ahmed, F.: Diffusion models beat gans on topology optimization. In: Proceedings of the AAAI Conference on Artificial Intelligence (AAAI), Washington, DC (2023)

\bibitem{muller2022diffusion}
M{\"u}ller-Franzes, G., Niehues, J.M., Khader, F., Arasteh, S.T., Haarburger, C., Kuhl, C., Wang, T., Han, T., Nebelung, S., Kather, J.N., et~al.: Diffusion probabilistic models beat gans on medical images. arXiv preprint arXiv:2212.07501  (2022)

\bibitem{ouyang2022training}
Ouyang, L., Wu, J., Jiang, X., Almeida, D., Wainwright, C., Mishkin, P., Zhang, C., Agarwal, S., Slama, K., Ray, A., et~al.: Training language models to follow instructions with human feedback. Advances in Neural Information Processing Systems  \textbf{35},  27730--27744 (2022)

\bibitem{radford2021learning}
Radford, A., Kim, J.W., Hallacy, C., Ramesh, A., Goh, G., Agarwal, S., Sastry, G., Askell, A., Mishkin, P., Clark, J., et~al.: Learning transferable visual models from natural language supervision. In: International conference on machine learning. pp. 8748--8763. PMLR (2021)

\bibitem{radford2019language}
Radford, A., Wu, J., Child, R., Luan, D., Amodei, D., Sutskever, I., et~al.: Language models are unsupervised multitask learners. OpenAI blog  \textbf{1}(8), ~9 (2019)

\bibitem{rombach2022high}
Rombach, R., Blattmann, A., Lorenz, D., Esser, P., Ommer, B.: High-resolution image synthesis with latent diffusion models. In: Proceedings of the IEEE/CVF conference on computer vision and pattern recognition. pp. 10684--10695 (2022)

\bibitem{ronneberger2015u}
Ronneberger, O., Fischer, P., Brox, T.: U-net: Convolutional networks for biomedical image segmentation. In: Medical Image Computing and Computer-Assisted Intervention--MICCAI 2015: 18th International Conference, Munich, Germany, October 5-9, 2015, Proceedings, Part III 18. pp. 234--241. Springer (2015)

\bibitem{ruiz2023dreambooth}
Ruiz, N., Li, Y., Jampani, V., Pritch, Y., Rubinstein, M., Aberman, K.: Dreambooth: Fine tuning text-to-image diffusion models for subject-driven generation. In: Proceedings of the IEEE/CVF Conference on Computer Vision and Pattern Recognition. pp. 22500--22510 (2023)

\bibitem{salem2019ml}
Salem, A., Zhang, Y., Humbert, M., Fritz, M., Backes, M.: Ml-leaks: Model and data independent membership inference attacks and defenses on machine learning models. In: Network and Distributed Systems Security Symposium 2019. Internet Society (2019)

\bibitem{schuhmann2022laion}
Schuhmann, C., Beaumont, R., Vencu, R., Gordon, C., Wightman, R., Cherti, M., Coombes, T., Katta, A., Mullis, C., Wortsman, M., et~al.: Laion-5b: An open large-scale dataset for training next generation image-text models. Advances in Neural Information Processing Systems  \textbf{35},  25278--25294 (2022)

\bibitem{schuhmann2021laion}
Schuhmann, C., Kaczmarczyk, R., Komatsuzaki, A., Katta, A., Vencu, R., Beaumont, R., Jitsev, J., Coombes, T., Mullis, C.: Laion-400m: Open dataset of clip-filtered 400 million image-text pairs. In: NeurIPS Workshop Datacentric AI. No. FZJ-2022-00923, J{\"u}lich Supercomputing Center (2021)

\bibitem{shokri2017membership}
Shokri, R., Stronati, M., Song, C., Shmatikov, V.: Membership inference attacks against machine learning models. In: 2017 IEEE symposium on security and privacy (SP). pp. 3--18. IEEE (2017)

\bibitem{song2020denoising}
Song, J., Meng, C., Ermon, S.: Denoising diffusion implicit models. In: International Conference on Learning Representations (2020)

\bibitem{song2020score}
Song, Y., Sohl-Dickstein, J., Kingma, D.P., Kumar, A., Ermon, S., Poole, B.: Score-based generative modeling through stochastic differential equations. In: International Conference on Learning Representations (2020)

\bibitem{swets1988measuring}
Swets, J.A.: Measuring the accuracy of diagnostic systems. Science  \textbf{240}(4857),  1285--1293 (1988)

\bibitem{truex2019demystifying}
Truex, S., Liu, L., Gursoy, M.E., Yu, L., Wei, W.: Demystifying membership inference attacks in machine learning as a service. IEEE Transactions on Services Computing  \textbf{14}(6),  2073--2089 (2019)

\bibitem{van2020ethical}
Van~Noorden, R.: The ethical questions that haunt facial-recognition research. Nature  \textbf{587}(7834),  354--359 (2020)

\bibitem{wu2023tune}
Wu, J.Z., Ge, Y., Wang, X., Lei, S.W., Gu, Y., Shi, Y., Hsu, W., Shan, Y., Qie, X., Shou, M.Z.: Tune-a-video: One-shot tuning of image diffusion models for text-to-video generation. In: Proceedings of the IEEE/CVF International Conference on Computer Vision. pp. 7623--7633 (2023)

\end{thebibliography}
\end{document}